\pdfoutput=1
\documentclass[11pt]{article}

\usepackage{ACL2023}
\usepackage{caption}
\usepackage{tcolorbox}
\usepackage{listings}
\usepackage{tabularx} 
\usepackage{times}
\usepackage{ragged2e}
\usepackage{latexsym}
\usepackage{svg}
\usepackage{graphicx}
\usepackage{array}
\usepackage{hyperref}
\usepackage[normalem]{ulem}
\usepackage[T1]{fontenc}
\usepackage{mathtools}
\usepackage{adjustbox}
\usepackage{diagbox}
\usepackage{float} 

\usepackage{xcolor}
\usepackage{soul}
\definecolor{MyTurquoise}{RGB}{175,238,238} 
\definecolor{MyRed}{RGB}{255,204,203}
\definecolor{MyYellow}{RGB}{255,255,102}

\usepackage[utf8]{inputenc}

\usepackage{microtype}

\usepackage{inconsolata}
\usepackage[shortlabels]{enumitem}
\usepackage{tabularx, booktabs, subcaption, multirow}
\usepackage{longtable, multicol}
\title{LLM for Complex Reasoning Task:An Exploratory Study in Fermi Problems}

\author{Zishou Liu\textsuperscript{1}, Carlos Rabat Villarreal\textsuperscript{2}, Mostafa Rahgouy\textsuperscript{2}, Amit Das \textsuperscript{3}, Zheng Zhang\textsuperscript{4}\\ \textbf{Chang Ren}\textsuperscript{2},   
\textbf{Dongji Feng}\textsuperscript{1},\\
\textsuperscript{1}  MCS department, Gustavus Adolphus College
\\
\textsuperscript{2} Department of CSSE, Auburn University\\
\textsuperscript{3} Department of CIS, University of North Alabama
\\
\textsuperscript{4} Department of CSIS, Murray State University\\
\{tomliu, djfeng\}@gustavus.edu}

\begin{document}
\maketitle

\begin{abstract}

Fermi Problems (FPs) are mathematical reasoning tasks that require human-like logic and numerical reasoning. Unlike other reasoning questions, FPs often involve real-world impracticalities or ambiguous concepts, making them challenging even for humans to solve. Despite advancements in AI, particularly with large language models (LLMs) in various reasoning tasks, FPs remain relatively under-explored.
This work conducted an exploratory study to examine the capabilities and limitations of LLMs in solving FPs. We first evaluated the overall performance of three advanced LLMs using a publicly available FP dataset. We designed prompts according to the recently proposed TELeR taxonomy, including a zero-shot scenario. Results indicated that all three LLMs achieved a fp\_score (range between 0 - 1) below 0.5, underscoring the inherent difficulty of these reasoning tasks. 
To further investigate, we categorized FPs into standard and specific questions, hypothesizing that LLMs would perform better on standard questions, which are characterized by clarity and conciseness, than on specific ones.
Comparative experiments confirmed this hypothesis, demonstrating that LLMs performed better on standard FPs in terms of both accuracy and efficiency. The paper related code has been published on: 
 \url{https://github.com/Gusties-stu/LLM\_FPs\_Explortory.git}

\end{abstract}

\section{Introduction}\label{sec:intro}

\begin{figure*}[!htb]
    \centering
    \includegraphics[width=0.8\linewidth]{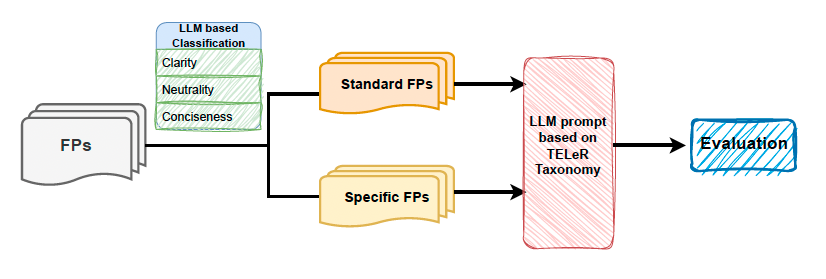}
    \vspace{-4mm}
    \caption{LLM boosted Pipeline to Solve Fermi Reasoning Problems}
    \label{fig:prompts_pipeline}
    \vspace{-2mm}
\end{figure*}


Reasoning in human behavior involves using logic and prior knowledge to make predictions, draw conclusions, or provide explanations~\cite{manktelow2012thinking}.
Mathematical reasoning is a crucial aspect of human intelligence, allowing us to comprehend and make decisions based on numerical data and language~\cite{lu2022survey}. Recent advances in employing Large Language Models (LLMs)~\cite{brown2020language,touvron2023llama,scao2022bloom,chowdhery2022palm} on various reasoning tasks have demonstrated impressive performance. To better understand the reasoning capabilities of LLMs, previous researchers have explored either the architecture side (including the addition of an extra evaluator based on Synthetic Data)~\cite{yoran2023answering,yin2024reasoning,lu2024mathgenie,tong2024can} or prompt design side ~\cite{wei2022chain,wang2022self,yao2024tree}.
However, the complexity and inherent characteristics of individual questions in these studies have not been thoroughly investigated.

To address this issue, we explored Fermi Problem (FPs), a challenging math reasoning task that is relatively understudied ~\cite{yoran2023making, kalyan2021much,rahgouy2023navigating}. FPs are a type of complex problems that require numerical reasoning but often lack precise answers due to impracticality or ambiguous concepts. For instance, in a typical FP, \textit{How much would the ocean surface rise if the ice caps melted}, different resources may have varying definitions of ``ocean surface''.  Additionally, FPs can include private or professional terms that LLMs may not have encountered before, making it difficult for LLMs to understand vague concepts and perform well on these questions~\cite{li2024personal}. While fine-tuning LLMs for private environments is an option, the cost and time involved can be significant in many cases. ~\cite{zhang2024scaling,xia2024understanding}
To better understand the behavior of LLMs on complex real-world reasoning tasks, we evaluated three advanced LLM models (GPT\textit{3.5}/\textit{4} and Llama\textit{3}) to solve the real FP dataset~\cite{kalyan2021much},  designing our prompts according to the recently introduced TELeR taxonomy~\cite{santu2023teler}, including one zero-shot prompt. Surprisingly, despite detailed and structured prompt engineering, all three LLMs achieved fp\_score below 0.5 (on a scale from 0 to 1).
The following question is 
how can we analyze FPs to uncover their inherent properties and assess whether LLMs interpret them differently?
We begin by categorizing the FPs into different types of questions. One approach is to classify questions based on certain characteristics. To simplify the task, we consider a binary classification and utilize the following three features: \textit{clarity}, \textit{neutrality}, and \textit{conciseness}.(see Section \ref{sec:Motivation} for detailed definition). If an FP meets these criteria, we classify it as a ``Standard Question'', meaning that its semantic meaning and formulation are clear and unambiguous. Otherwise, we classify it as a ``Specific Question'', indicating that this FP is case-specific, rarely accessible, or has unclear meaning.

We tested our classification on the same FP dataset ~\cite{kalyan2021much} based on our proposed features.
After the classification, we have two sets of FPs. Next, we conduct a comprehensive study of the FPs using three LLMs on the two sets of FPs and compare their performance. Thus, we created an LLM-based end-to-end pipeline to solve FPs as shown in Figure~\ref{fig:prompts_pipeline}.


In summary, we make the following contributions to this paper:

\begin{enumerate}[leftmargin=*,itemsep=0ex,partopsep=0ex,parsep=0ex]

\item We developed a pipeline solely using LLMs to solve a real-world reasoning challenge: Fermi Problems

\item  We defined an intuitive  method to classify FPs and made an exploring investigation while conducting three LLMs

\item Our results demonstrate varying LLM performance on different types of FPs, providing insights into LLM understanding of real-world problems.

\end{enumerate}

\section{Related Work}\label{sec:related}


\noindent\textbf{Fermi Problems:} 
Fermi Problems (FPs) are a class of real-world problems that require numerical reasoning within the scope of human intelligence
~\cite{lu2022survey}.
These problems, which we frequently encounter in everyday life, pose questions that can only be approximately estimated due to the impracticality or impossibility of precise computation. For example, an NLP researcher might be curious about:\textit{ How much coffee was consumed during ACL 2025?}
FPs were first introduced as an NLP task by~\cite{kalyan2021much}.
Recently, FPs have gained attention for evaluating the performance of different Large LLMs and various prompt designs. For instance, ~\cite{yoran2023making}  trained retrieval-augmented language models on four QA tasks, discovering that FPs were the most challenging, yielding the lowest performance.
Similarly, ~\cite{rahgouy2023navigating}  used FPs to assess LLMs' capabilities in handling Complex Multi-hop Queries, testing models across fine-tuning, few-shot/zero-shot learning, and Chain-of-Thought (CoT) prompting techniques on four different LLMs.
Meanwhile, ~\cite{yoran2023answering} introduced Multi-Chain Reasoning (MCR), a method that prompts LLMs to meta-reason over multiple CoTs rather than aggregating their answers. They tested MCR on seven multi-hop QA datasets, including FPs, and consistently found FPs to be the most challenging task.
\noindent\textbf{Reasoning challenge:} 
Reasoning, a fundamental cognitive process integral to human intelligence, has gained significant interest within the AI community~\cite{lu2022survey}. Mathematical reasoning is one of the primary challenges in this domain.  Specifically, NLP researchers defined Math word problem solving (MWPs) ~\cite{zhao2020ape210k, cobbe2021training} where MWPs (also known as algebraic or arithmetic word problems) present a brief narrative involving characters, entities, and quantities, with solutions that are typically numerical and explicit~\cite{lu2022survey}. In addition to MWPs, researchers have explored other types of reasoning tasks such as Theorem Proving~\cite{polu2020generative,han2021proof,polu2022formal} , Geometry Problem solving ~\cite{Chen2021GeoQAAG,Cao2022AnAB}. Although FPs belong to the broader category of Math Question Answering (MathQA), they pose a unique challenge due to the approximate nature of their solutions. This increased difficulty arises from the inability to precisely calculate the answers, making FPs more demanding than traditional MWPs.

\noindent\textbf{Prompt Design:} 
A prompt for LLM is a collection of guidelines that can direct the Language Model toward a particular task. ~\cite{liu2023pre}.
Currently, different prompt design strategies are proposed to improve the performance of LLM in terms of such reasoning and decision-making ability ~\cite{zhou2022least, wei-2022-entropy,yao-etal-2021-adapt}. The response of LLMs can vary significantly based on the prompt's quality. This variation can be attributed to different LLMs' diverse training data-sets and annotations. Therefore, utilizing the same prompt to evaluate/explain multiple LLMs is advisable to facilitate a meaningful comparison of the performances of various LLMs ~\cite{santu2023teler}.

\noindent\textbf{Distinct from previous work:} Although substantial Question Answer (QA) research has focused on reasoning about multiple facts (whether relevant or distracting), many approaches rely on augmenting LLMs with additional retrieval models, which requires significant effort. 

Meanwhile, research has not systematically explored the characteristics of FPs, nor has it developed customized implementations to address them.
Our work aims to investigate the implications of using intuitive criteria to classify FPs and evaluate LLMs' performance across various FP types, utilizing prompts designed based on the TELeR taxonomy.

\section{Research Objective}\label{sec:Motivation}

Upon closer examination of FPs, it becomes evident that these questions generally encompass two fundamental types of information: standard information and specific information. The following examples illustrate this distinction:

\textbf{Standard question:} "What fraction of the \textit{sun}'s energy output is intercepted by the \textit{Earth}?"

\textbf{Specific question:} "How many golf balls put into the \textit{world's oceans }would it take to submerge all of the land on Earth from the displaced water?"

In the first case, concepts such as the \textit{sun} and the \textit{earth} are standard and widely recognized across various sources. Conversely, the concept of \textit{world's oceans} in the second case has various definitions within a specific case, which suggest that obtaining standardized and precise data or consistent information in the training data can be challenging.

We argue that although a significant portion of the data used to train state-of-the-art LLMs is derived from publicly available internet resources ~\cite{kojima2022large,51308}, the presence of unavailable or inconsistent information can negatively impact the performance of LLMs across different domains. This discrepancy leads to better performance on standard questions compared to specific ones.
Thus, we conclude our first hypothesis:

\textit{Hypothesis: LLMs perform better on standard questions compared to specific questions due to the presence of unclear or inconsistent concepts within the latter.
}

The following question is: how can this issue be addressed while minimizing the effort required? We propose leveraging the recently published prompt taxonomy to design and test prompts specifically tailored for FPs.
To better support our assumption, we first define the following three features that are commonly associated with standard questions:

\begin{enumerate}
    \item \textbf{Clarity}:
    The standard questions should have at least one intended meaning and be easy to understand without obvious ambiguity.
    \vspace{-2mm}
    \item \textbf{Neutrality}:
    The standard questions should not be biased towards any particular viewpoint.
    \vspace{-2mm}
    \item \textbf{Conciseness}:
    The standard questions should be concise and to the point.
\end{enumerate}


While the other questions that do not meet these conditions will be classified as \textbf{specific questions} in this binary classification. We then utilize GPT-turbo-3.5 to classify the FPs based on our criteria. The prompt for LLMs to do classification  can be seen in Appendix \ref{sec:template2}. We selected GPT-3.5 due to its balanced classification performance, particularly in distinguishing between standard and specific categories, whereas other models exhibit more uneven distributions. Also, the reason we only use one LLM to do classification is that we want to stick to one method to maintain consistency on the working pipeline. Investigating the utility of different classifiers is not the goal of this paper and will be covered in future work. Statistical information for two types of questions is in Table \ref{tab:Question_Classification}.

\begin{table}[!hbt] 
\centering
\small 
\setlength{\tabcolsep}{0.6pt} 
\renewcommand{\arraystretch}{0.9} 
\begin{tabular}{|c|p{2cm}|p{2cm}|}
  \hline
  \diagbox{Model Name}{Question Type} & Standard & Specific \\
  \hline
  GPT-turbo-3.5 &   252 &   304  \\
  \hline
\end{tabular}
\caption{Classification of Fermi Problem question types using GPT-turbo-3.5 as the classifier.}
\label{tab:Question_Classification}
\end{table}

After selection, the final result of our dataset included 252 standard questions (45.32\%) and 304 specific questions (54.68\%). We only used 556 samples from the original 558 testing cases because we found that 2 samples had an inaccurate gold standard.








\section{Background}\label{sec:Fermi question}

\subsection{Fermi Problem}

 The Fermi challenge is inspired by the Nobel Prize-winning physicist Enrico Fermi, is a fascinating exercise in estimation often referred to as “Fermi problems” (FPs) \cite{kalyan2021much}. Fermi was renowned for his extraordinary ability to make accurate estimates for complex numerical problems using minimal data. These problems are not only about arriving at precise calculations but rather involve making reasoned assumptions and approximations to reach a rough estimate, thus better aligned with real-world application scenarios.

FPs typically address issues where solutions are either too challenging to measure directly or inherently imprecise.  For example, a typical FP would be:"\textit{How much would the ocean surface rise if the ice caps melted}."
Specifically, FPs require human-like intelligence such as 1) Mathematical reasoning, 2) Question Decomposition, 3) Common sense, 4) Numerical Estimation.


\subsection{Task Description}
Previous researchers ~\cite{kalyan2021much, yoran2023making, rahgouy2023navigating} approached FPs by breaking them down into three progressively complex tasks. The simplest of these tasks involves providing only the relevant facts, denoted as \textit{F}, alongside the question \textit{Q}. This setup is then extended in Task 2, where a set of distract facts is introduced, requiring the model to discern which facts are pertinent to the solution.

Our work focuses on Task 3, which represents the original and most challenging FP setting. In this scenario, the input consists solely of the questions, without any provided facts or distracts. Due to the unconstrained nature of this task, arriving at a precise answer for FPs is inherently difficult. As such, in the Fermi Science Olympiads, participants are awarded full points for answers that fall within the same order of magnitude as a reference gold answer. If their answers deviate by an order of magnitude, they receive 1/3 fewer points.
For further details on the evaluation process, refer to~\ref{sec:eval}.



\section{Experiment Design}\label{sec:Experiment}





\subsection{Dataset}

We used the REALFP dataset\footnote{https://allenai.org/data/fermi}
 from ~\cite{kalyan2021much} which contains 185, 185, and 558 questions, respectively, for training, validation, and testing. These questions cover a wide range of topics, requiring domain-specific reasoning such as physics, basic mechanics of Poker, etc. 


\subsection{Evaluation}\label{sec:eval}

FPs typically address issues where solutions are either too challenging to measure directly or inherently imprecise. We utilize the \textbf{fp\_score}  evaluation metric from ~\cite{kalyan2021much}:

\begin{equation}
    \text{fp\_score} = \max\left(0, 1 - \frac{1}{3} \left| \log_{10}\left(\frac{A'}{A}\right) \right| \right) \tag{1}
\end{equation}

This metric is designed to capture the nuances of imprecision and uncertainty in model predictions \cite{rahgouy2023navigating}. It assigns a full score when the predicted answer aligns within the same order of magnitude as the gold standard reference, reflecting the model's proximity to the expected outcome. However, as the prediction drifts further from this reference specifically, with each order of magnitude the score is progressively reduced by one-third. Here, A' represents the output generated by the LLM model, while A denotes the gold standard value sourced from our dataset. The evaluation metric is normalized, ranging from 0 to 1, where 1 signifies perfect alignment and 0 indicates a complete divergence from the reference.

\subsection{Large Language Models}

For this experiment, we selected three of the most popular commercial large language models: OpenAI GPT(include GPT-3.5 and GPT-4) and Meta Llama. We utilized their respective APIs. Table ~\ref{tab:LLMs} gives more details about these three LLMs.

\begin{table}[!hb]\footnotesize
\centering
\setlength{\tabcolsep}{1pt} 
\begin{tabular}{|l|l|p{1.5cm}|p{3.5cm}|p{4cm}|}
\hline
\multicolumn{5}{|c|}{LLM Information} \\ \hline
\multicolumn{1}{|l|}{Abbreviation} & \multicolumn{2}{l|}{Model Full Name} & \multicolumn{2}{l|}{Model Information} \\ \hline
\multicolumn{1}{|l|}{GPT-3.5} & \multicolumn{2}{l|}{OpenAI GPT-3.5 Turbo} & \multicolumn{2}{l|}{Unknown Parameter} \\ \hline
\multicolumn{1}{|l|}{GPT-4} & \multicolumn{2}{l|}{OpenAI GPT-4 Turbo} & \multicolumn{2}{l|}{Unknown Parameter} \\ \hline
\multicolumn{1}{|l|}{Llama 3.0} & \multicolumn{2}{l|}{Llama 3.0} & \multicolumn{2}{l|}{70B parameters} \\ \hline
\end{tabular}
\caption{Large language models have been studied and used in this paper. The abbreviation name provided in Table will be used throughout this paper.  }
\label{tab:LLMs}
\end{table}

\subsection{Prompt Design with TELeR taxonomy}


In this work, we conducted the TELeR taxonomy ~
\cite{santu2023teler}, which has been used in many NLP tasks while prompting LLMs ~\cite{santu2024prompting, salvador2024benchmarking}. Specifically, TELeR taxonomy categorized complex task prompts based on the following four criteria: 1)Turn, 2) Expression, 3) Level of Details, and 4) Role. Particularly, we used prompt difficulty levels 0, 2, and 4.  Prompt level 0 considers LLM as a zero-shot learner, without providing any directive to LLMs. Prompt level 2 requires a structured input format. Additionally, level 4 further injects the expected evaluation direction. The TELeR details can be seen in Appendix~\ref{sec:appendix}.

Since each FPs is unique, we utilize LLMs to generate prompts based on the TELeR taxonomy and particular questions. To achieve this, we rely on LLMs to create prompts using precise definitions for each prompt level, eliminating the need for manual input. For example, when evaluating GPT-4, we first use GPT-4 to generate a prompt based on that FP, then employ the generated prompt and FP to produce the final results.
An example of our machine-generated prompt can be found in appendix \ref{sec:template3}.

\section{Results and Analysis}\label{sec:Result}



\subsection{Higher level prompts Improve the LLMs Performance}

Table \ref{tab:question_score_prompt} provides a comparative analysis of the overall fp\_score achieved by three different LLMs across three distinct prompt levels: Level 0, Level 2, and Level 4. The results indicate that the performance of each model varies depending on the prompt level, and GPT-4 consistently achieves higher scores as the prompt level increases, culminating in a fp\_score of 0.500 at Level 4. Similarly, Llama shows a gradual improvement, although the increments are slightly smaller compared to GPT-4. Interestingly, GPT-3.5 demonstrates less variation across the prompt levels, suggesting a more stable but slightly lower performance across different levels of prompt complexity.Overall, we observe that all fp\_scores are below 0.5, indicating the difficulty LLMs have in solving FPs.

\begin{table}[!hbt]
\centering
\small 
\setlength{\tabcolsep}{1pt} 
\renewcommand{\arraystretch}{0.9} 
\begin{tabular}{|c|c|c|c|}
  \hline
  \diagbox{Model Name}{Prompt Level} & Level 0 & Level 2 & Level 4 \\
  \hline
  GPT-3.5 & 0.398 & 0.420 & 0.418 \\
  \hline
  GPT-4.0 & 0.466 & 0.479 & 0.500 \\
  \hline
  Llama 3.0 & 0.439 & 0.469 & 0.460 \\
  \hline
\end{tabular}
\caption{The overall fp\_score for different models based on three level prompts}
\label{tab:question_score_prompt}
\end{table}

\begin{table}[!hbt]
\centering
\resizebox{\columnwidth}{!}{%
\begin{tabular}{|l|l|l|l|l|}
\hline
Model & Prompt Level & Standard  & Specific  & Gap \\ \hline
\multicolumn{1}{|c|}{\multirow{3}{*}{GPT-3.5}} & 0 & \textbf{0.442} & 0.3674 & 0.0746 \\ \cline{2-5} 
\multicolumn{1}{|c|}{} & 2 & \textbf{0.491} & 0.3671 & 0.1239 \\ \cline{2-5} 
\multicolumn{1}{|c|}{} & 4 & \textbf{0.500} & 0.353 & 0.1470 \\ \hline
\multirow{3}{*}{ GPT-4} & 0 & \textbf{0.511} & 0.467 & 0.0440 \\ \cline{2-5} 
 & 2 & \textbf{0.527} & 0.481 & 0.0460 \\ \cline{2-5} 
 & 4 & \textbf{0.568} & 0.499 & 0.0690 \\ \hline
\multirow{3}{*}{ Llama 3.0} & 0 & \textbf{0.5455} & 0.439 & 0.1065 \\ \cline{2-5} 
 & 2 & \textbf{0.5458} & 0.468 & 0.0778 \\ \cline{2-5} 
 & 4 & \textbf{0.556} & 0.459 & 0.0970 \\ \hline
\end{tabular}%
}
\caption{The fp\_score of LLMs on Standard Vs. Specific question types across prompt levels, with corresponding gap values.}
\label{tab:prompts_taxonomy_score}
\end{table}

\subsection{LLM prefers Standard Questions than Specific Questions}
We hypothesize that LLMs should perform better on standard questions than on specific questions, owing to the inconsistent and unclear concepts often present in the latter. Based on our classification of standard Vs.  specific questions (see Section ~\ref{sec:Motivation}
), we sampled 252 standard questions (45.32\%) and 304 specific questions (54.68\%) from the testing set.

Table \ref{tab:prompts_taxonomy_score} summarizes the detailed fp\_score  of how three LLMs respond to varying levels of prompt structure when tackling standard Vs. specific questions. 
We first compare the performance of GPT family on two question sets. For example, at prompt level 0, GPT-3.5 achieves a score of 0.442 on standard questions and 0.3674 on specific questions. This indicates a performance gap of approximately 0.0746, showing that GPT-3.5 handles standard questions better. Interestingly, when prompt level increases to 4, the standard question score rises slightly to 0.500, while the specific question score drops to 0.353. The gap of performance widens to 0.147, further emphasizing the model’s stronger performance on standard questions as prompt complexity increases.  

For GPT-4, on the baseline (prompt level 0), GPT-4 scores 0.511 on standard questions and 0.467 on specific questions, with a gap of 0.044. Although it is smaller than GPT-3.5, the gap still indicates better performance on standard questions. With prompt level 4, GPT-4's standard question score increases to 0.568, and the specific question score increases to 0.499, widening the gap to 0.069. This demonstrates that while GPT-4's performance improves overall with prompt structure, the gap between standard and specific question performance remains, suggesting a persistent challenge in handling specific questions.

For Llama, at prompt level 0, Llama has a standard question score of 0.5455 and a specific question score of 0.439, resulting in a significant gap of 0.1065 indicating Llama's greater struggle with specific questions. When the prompt level is increased to 4, the standard question score increases slightly to 0.556, while the specific question score reaches 0.459. The gap narrows to 0.097, showing the model still performs noticeably better on standard questions.

Across all models and prompt levels, standard questions consistently yield higher scores compared to specific questions. The gaps of performances indicate that specific questions are more challenging for the LLMs, highlighting the inherent difficulty these models face in generating precise, contextually relevant answers when the task requires detailed specificity.

\subsection{Deep Investigation on Higher Prompt Levels}

In this section, we examine three perspectives on the impact of TELeR prompt taxonomy on LLMs.

\subsubsection{High Prompt Level Increase the Distinguishability of Metric Score}

This first investigation focuses on the distinguishability of fp\_score when using the TELeR prompt taxonomy.   This perspective is to quantify whether the evaluation metric (fp\_score in this case) is able to distinguish between different question types. Also, we will discuss the further utility of this purpose in Conclusion and Future Work Section ~\ref{sec:conclusion}.

We utilized a metric to quantify \textit{percentage absolute difference (PAD)}~\cite{feng2023joint}. This metric was originally used to quantify the distinguishing ability of an evaluation metric in terms of two classification methods. The higher distinguishability will result in higher \textit{PAD} between pair of ranking methods. Mathematically, we use the following formula for  \textit{PAD} between Standard Vs. Specific questions in terms of their fp\_scores:

\vspace{-2mm}
{
\begin{equation}\tag{2} \label{equ:pad}\small
    PAD^{}_{PL}=\frac{|fp\_score^{Standard}_{PL} -fp\_score^{Specific}_{PL}|}{ \max\left(fp\_score^{Standard}_{PL},fp\_score^{Specific}_{PL}\right)} 
\end{equation}
}


Here, $PL$ represents a specific "prompt level" of fp\_score achieved either from a ``standard question'' or ``specific question''.  From Table ~\ref{tab:PAD}, we can observe that while using TELeR prompt taxonomy, the \textit{PAD} score of level 4 is higher than the other prompt in the GPT family (GPT-3.5 and GPT-4). For instance, the \textit{PAD} score of prompt 4 in GPT-3.5 is \textit{0.294}, while for GPT-4, prompt level 4 still achieved the highest which is \textit{0.121}. However, in Llama, this number decreased slightly from 0.195 to 0.189, reflecting a minor reduction in the usefulness of the prompt level for distinguishability.

\begin{table}[!hbt]
\centering
\small 

\scalebox{0.98}{
\begin{tabular}{|l|l|ll|}
\hline
Model & Prompt Level & \multicolumn{2}{l|}{PAD} \\ \hline
\multicolumn{1}{|c|}{\multirow{3}{*}{GPT-3.5}} & 0 & \multicolumn{2}{l|}{0.168} \\ \cline{2-4} 
\multicolumn{1}{|c|}{} & 2 & \multicolumn{2}{l|}{0.252} \\ \cline{2-4} 
\multicolumn{1}{|c|}{} & 4 & \multicolumn{2}{l|}{\textbf{0.294}} \\ \hline
\multirow{3}{*}{ GPT-4} & 0 & \multicolumn{2}{l|}{0.086} \\ \cline{2-4} 
 & 2 & \multicolumn{2}{l|}{0.087} \\ \cline{2-4} 
 & 4 & \multicolumn{2}{l|}{\textbf{0.121}} \\ \hline
\multirow{3}{*}{ Llama 3.0} & 0 & \multicolumn{2}{l|}{0.195} \\ \cline{2-4} 
 & 2 & \multicolumn{2}{l|}{0.142} \\ \cline{2-4} 
 & 4 & \multicolumn{2}{l|}{0.189} \\ \hline
\end{tabular}%

}

\caption{Percentage Absolute Difference between Standard Versus Specific Questions in terms of different levels of prompts.}
\label{tab:PAD}
\end{table}

\subsubsection{High Prompt Level Improve Efficiency of LLMs}

Next, we evaluate the efficiency of LLMs across different prompting levels. Previous researchers use ``multi-hop'' prompts to enhance reasoning performance; though the better results, additional hops also lead to increased token usage and higher computational costs ~\cite{biran2024hopping,yang2024large}.
Fewer hops indicate a more responsive and efficient model, while a higher number of hops suggests a slower reaction, reflecting weaker comprehension and answering ability for tasks.

We test the number of hops required to solve Standard Vs. Specific questions while using different prompt levels. 
When prompted to solve a FP, the LLM generates a paragraph containing numerical answers as the final result. To evaluate correctness, we detect and extract these numerical results. If no numerical value is found, the prompt is iterated until a valid result is produced. To avoid infinite loops, we set a maximum of 10 attempts per prompt for each LLM.

Table ~\ref{tab:Attempts_Number_Stats} shows the result and concludes the number of hops that LLMs require to generate valid results across various prompt levels for both standard and specific questions. In particular, if an LLM generates an answer that is entirely nonsensical (i.e., without a numeric value indicating a feasible FP result), we incremented the number of hops required for this FP example.
The same prompt will be used to generate a new response from the model until a calculable answer is achieved.

Note that a "calculable answer" does not mean a correct answer, but a valid answer that we can extract the numeric value. In this Table 
\ref{tab:Attempts_Number_Stats}, "\textbf{Total number of Hop}'' represents the total number of extra hops we need to prompt LLM to provide a calculable answer for the entire datasets. "\textbf{Samples more than one hop}" shows the number of samples that require more than one hop to achieve a calculable answer. Note that the minimum number of hops for each sample is 1. 

Based on the presented data, it is evident that LLMs solve standard questions more efficiently than specific questions.
For example, solving all standard questions required a total of 260 hops across three LLMs, whereas solving specific questions required 354 hops. Additionally, the number of samples need more than one hop was lower for standard questions (83) compared to specific questions (95).
We can also observe that 
prompt level 4 can decrease the number of hops required for the GPT family in both standard questions and specific questions. For instance, GPT-4 requires 92 (82) hops while solving all standard questions (specific questions) using prompt level 0, these two numbers dropped to 38 (69) while using prompt level 4, which is an interesting observation. 
For Llama, neither standard nor specific requires more hops, indicating that Llama is better suited to achieve valid numeric results, compared to the GPT family.


\begin{table}[!hbt]
\centering
\resizebox{\columnwidth}{!}{%
\begin{tabular}{|l|l|l|l|l|l|}
\hline
Model & Prompt Level & \multicolumn{2}{c|}{Total \# of hop} & \multicolumn{2}{c|}{\begin{tabular}[c]{@{}l@{}}Samples more\\ than one hop\end{tabular}} \\ \hline
 &  & Standard & Specific & Standard & Specific \\ \hline
\multicolumn{1}{|c|}{\multirow{3}{*}{GPT-3.5}} & 0 & \textbf{27} & 33 & \textbf{11} & 8 \\ \cline{2-6}
\multicolumn{1}{|c|}{} & 2 & \textbf{55} & 54 & \textbf{22} & 20 \\ \cline{2-6}
\multicolumn{1}{|c|}{} & 4 & \textbf{15} & 18 & \textbf{7} & 7 \\ \hline
\multirow{3}{*}{ GPT-4} & 0 & \textbf{92} & 82 & \textbf{24} & 17 \\ \cline{2-6}& 2 & \textbf{30} & 89 & \textbf{10} & 24 \\ \cline{2-6}& 4 & \textbf{38} & 69 & \textbf{6} & 14 \\ \hline
\multirow{3}{*}{ Llama 3.0} & 0 & \textbf{2} & 2 & \textbf{2} & 2 \\ \cline{2-6}& 2 & \textbf{0} & 6 & \textbf{0} & 2 \\ \cline{2-6}& 4 & \textbf{1} & 1 & \textbf{1} & 1\\ \hline
\textbf{Sum} & & \textbf{260} & \textbf{354} & \textbf{83} & \textbf{95} \\ \hline
\end{tabular}%
}
\caption{Comparison of total hops and multi-hop samples used by LLMs for standard versus specific questions.}
\label{tab:Attempts_Number_Stats}
\end{table}

\vspace{-2mm}

\subsubsection{High Prompt Level Improve Consistency of LLMs}
\begin{figure*}[!hbt]
    \centering
    \includegraphics[width=0.85\linewidth, height=1.0\textheight, keepaspectratio]{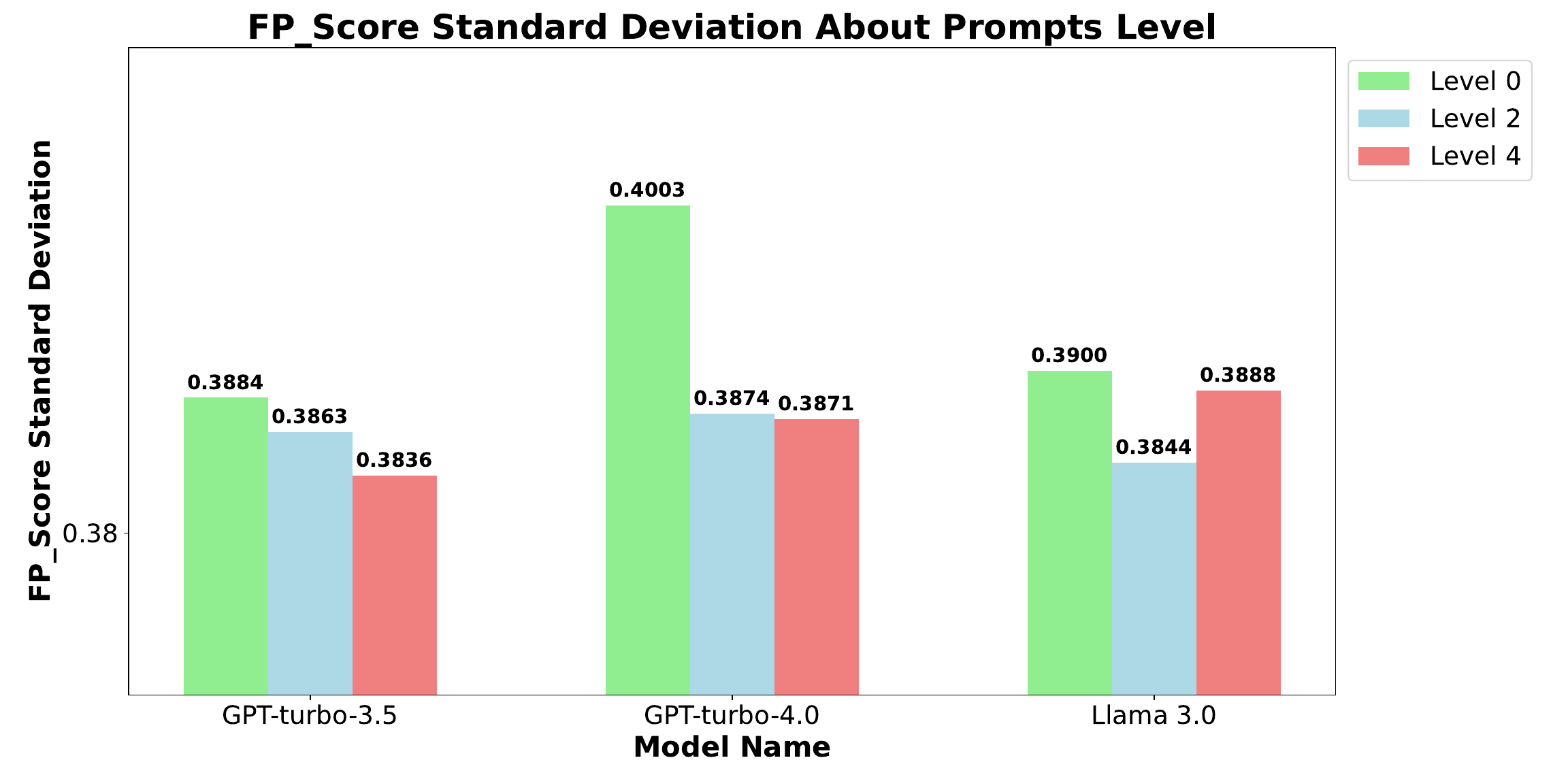}
    \caption{Standard deviation on FPs on different models with scores of different prompt level .}
    \label{fig:std_dev_prompt_levels}
\end{figure*}

Our third perspective analyzes the consistency of LLM performance while using different prompt levels. 
In Figure ~\ref{fig:std_dev_prompt_levels}, we show the standard deviation of fp\_scores achieved at various prompt levels for all types of questions. A lower standard deviation indicates greater stability and consistency between different FPs.

As demonstrated by the results, the effectiveness of prompt design in achieving consistent outcomes is evident. For instance, within the GPT family, prompt level 4 achieved the lowest standard deviation across different FPs questions, reflecting the most stable performance. Notably, without any prompt injection (prompt level 0), GPT-4 exhibited the highest standard deviation (0.4003) among all LLMs, indicating the least stable performance. However, prompt level 4 reduced this deviation to 0.387, resulting in a 0.03\% improvement in stability.
Interestingly, for the Llama model, prompt level 2 achieved the lowest standard deviation, highlighting a notable divergence in optimal prompt levels between different LLMs.

\subsection{Deep Investigation on Question Type}

From our previous experiments, we observed that LLMs exhibit different capabilities when responding to standard questions compared to specific questions. In this section, we delve deeper into the reasons behind this difference. As is well known, LLMs generate text by predicting the probability distribution of the next token (word or sub words) based on the preceding context, with embedding playing a crucial role in this process~\cite{vaswani2017attention, radford2019language, brown2020language}. 
A key factor influencing LLM performance on specific questions is how the model understands and processes the embedding of vague or contextually complex words, which can significantly affect its comprehension of such questions.

To ensure consistency in our analysis, we employed GPT-3.5 to classify the question types. We also utilized the \textit{OpenAI text-embedding-ada-002 model}
\footnote{https://openai.com} to capture the semantic embedding information.
Using this model, we calculated: 

1) the average cosine similarity between standard questions. 
2) the average cosine similarity between specific questions. 
3) the average cross-similarity between standard and specific questions. 

Table ~\ref{tab:embedding} presents the similarity scores for these three types of question sets. As observed, the similarity within specific types of questions—whether standard (0.1736) or specific (0.1821)—is higher than the cross-similarity between standard and specific questions (0.1653). This result indicates a notable difference in how the model understands embedding for these different types of questions.

\begin{table}[!htb] 
\centering
\small 
\setlength{\tabcolsep}{4pt}
\begin{tabular}{|c|c|}
  \hline
  {Question Type} & Average Score \\
  \hline
  Standard Vs. Standard & 0.1736 \\
  \hline
  Specific Vs. Specific & 0.1821 \\
  \hline
  Mix & 0.1653 \\
  \hline
\end{tabular}
\caption{The cosine similarity average scores for embedding of the three type questions group.}
\label{tab:embedding}
\end{table}

\section{Conclusion and Future Work}\label{sec:conclusion}

In this work, we systematically test three LLMs on FPs with different prompt levels based on the TELeR taxonomy. To achieve this, we created an LLM-based end-to-end pipeline, firstly using LLM to classify FPs into two types of questions: Standard Questions and Specific Questions, based on three intuitive criteria: \textbf{clarity}, \textbf{neutrality} and \textbf{conciseness}. After obtaining the two FP datasets, we conducted extensive experiments to analyze the accuracy and efficiency of LLMs w.r.t. two types of FP datasets.  Our work is summarized briefly as follows. 

\begin{enumerate}[leftmargin=*,itemsep=0ex,partopsep=0.5ex,parsep=0ex]

\item Our developed pipeline increases the 
performance of LLMs when using higher-level prompts to solve FPs in general.

\item LLMs perform better on standard questions compared to specific questions while using different levels of prompts.

\item High-level prompts significantly enhance the performance of LLMs by increasing LLM efficiency.

\end{enumerate}

Our work investigates the use of LLMs exclusively to create a pipeline to solve FPs. The proposed question classification method could be used to further analyze LLM performance in different types of questions, such as incorporating an additional classification step. Another potential direction is to introduce rewards/penalties based on the difficulty of question types, involving a weighted mechanism in the training process. 
\section{Limitations}\label{sec:limitation}

Our study acknowledges three key limitations:

\textbf{Dataset Size:} The limited dataset size of just 556 FPs constrain the generalizability of our findings. Currently, these 556 FPs constitute the only authorized and open-source dataset, as publicly available datasets on Fermi Questions remain scarce. Utilizing a larger and more diverse dataset in future work would allow for a more robust evaluation.  This limitation could be solved by collecting more FPs.

\textbf{Limited LLM Testing}: Given the budget, our evaluation was conducted on three LLMs. While this initial exploration demonstrates the efficacy of the TELeR taxonomy, including a broader range of LLMs in future studies, would provide a more comprehensive understanding of its generalizability.

\textbf{Implication of Standard Vs Specific Question}: Despite promising results, we acknowledge a limitation in our study regarding the interpretability of the short performance of LLMs on specific questions.  We specifically consider explaining performance from the perspective of attention mechanisms~\cite{bahdanau2014neural}. 
In addition, this study did not investigate the implications of these findings more deeply. As such, interpretability will be our dominant future direction.

\bibliography{anthology,custom}

\begin{thebibliography}{41}
\expandafter\ifx\csname natexlab\endcsname\relax\def\natexlab#1{#1}\fi

\bibitem[{Bahdanau et~al.(2014)Bahdanau, Cho, and Bengio}]{bahdanau2014neural}
Dzmitry Bahdanau, Kyunghyun Cho, and Yoshua Bengio. 2014.
\newblock Neural machine translation by jointly learning to align and translate.
\newblock \emph{arXiv preprint arXiv:1409.0473}.

\bibitem[{Biran et~al.(2024)Biran, Gottesman, Yang, Geva, and Globerson}]{biran2024hopping}
Eden Biran, Daniela Gottesman, Sohee Yang, Mor Geva, and Amir Globerson. 2024.
\newblock Hopping too late: Exploring the limitations of large language models on multi-hop queries.
\newblock \emph{arXiv preprint arXiv:2406.12775}.

\bibitem[{Brown et~al.(2020)Brown, Mann, Ryder, Subbiah, Kaplan, Dhariwal, Neelakantan, Shyam, Sastry, Askell et~al.}]{brown2020language}
Tom Brown, Benjamin Mann, Nick Ryder, Melanie Subbiah, Jared~D Kaplan, Prafulla Dhariwal, Arvind Neelakantan, Pranav Shyam, Girish Sastry, Amanda Askell, et~al. 2020.
\newblock Language models are few-shot learners.
\newblock \emph{Advances in neural information processing systems}, 33:1877--1901.

\bibitem[{Cao and Xiao(2022)}]{Cao2022AnAB}
Jie Cao and Jing Xiao. 2022.
\newblock \href {https://api.semanticscholar.org/CorpusID:252819519} {An augmented benchmark dataset for geometric question answering through dual parallel text encoding}.
\newblock In \emph{International Conference on Computational Linguistics}.

\bibitem[{Chen et~al.(2021)Chen, Tang, Qin, Liang, Liu, Xing, and Lin}]{Chen2021GeoQAAG}
Jiaqi Chen, Jianheng Tang, Jinghui Qin, Xiaodan Liang, Lingbo Liu, Eric~P. Xing, and Liang Lin. 2021.
\newblock \href {https://api.semanticscholar.org/CorpusID:235253782} {Geoqa: A geometric question answering benchmark towards multimodal numerical reasoning}.
\newblock \emph{ArXiv}, abs/2105.14517.

\bibitem[{Chowdhery et~al.(2022{\natexlab{a}})Chowdhery, Narang, Devlin, Bosma, Mishra, Roberts, Barham, Chung, Sutton, Gehrmann, Schuh, Shi, Tsvyashchenko, Maynez, Rao, Barnes, Tay, Shazeer, Prabhakaran, Reif, Du, Hutchinson, Pope, Bradbury, Austin, Isard, Gur-Ari, Yin, Duke, Levskaya, Ghemawat, Dev, Michalewski, Garcia, Misra, Robinson, Fedus, Zhou, Ippolito, Luan, Lim, Zoph, Spiridonov, Sepassi, Dohan, Agrawal, Omernick, Dai, Pillai, Pellat, Lewkowycz, Moreira, Child, Polozov, Lee, Zhou, Wang, Saeta, Diaz, Firat, Catasta, Wei, Meier-Hellstern, Eck, Dean, Petrov, and Fiedel}]{51308}
Aakanksha Chowdhery, Sharan Narang, Jacob Devlin, Maarten Bosma, Gaurav Mishra, Adam Roberts, Paul Barham, Hyung~Won Chung, Charles Sutton, Sebastian Gehrmann, Parker Schuh, Kensen Shi, Sasha Tsvyashchenko, Joshua Maynez, Abhishek Rao, Parker Barnes, Yi~Tay, Noam Shazeer, Vinodkumar Prabhakaran, Emily Reif, Nan Du, Ben Hutchinson, Reiner Pope, James Bradbury, Jacob Austin, Michael Isard, Guy Gur-Ari, Pengcheng Yin, Toju Duke, Anselm Levskaya, Sanjay Ghemawat, Sunipa Dev, Henryk Michalewski, Xavier Garcia, Vedant Misra, Kevin Robinson, Liam Fedus, Denny Zhou, Daphne Ippolito, David Luan, Hyeontaek Lim, Barret Zoph, Alexander Spiridonov, Ryan Sepassi, David Dohan, Shivani Agrawal, Mark Omernick, Andrew~M. Dai, Thanumalayan~Sankaranarayana Pillai, Marie Pellat, Aitor Lewkowycz, Erica Moreira, Rewon Child, Oleksandr Polozov, Katherine Lee, Zongwei Zhou, Xuezhi Wang, Brennan Saeta, Mark Diaz, Orhan Firat, Michele Catasta, Jason Wei, Kathy Meier-Hellstern, Douglas Eck, Jeff Dean, Slav Petrov, and Noah Fiedel.
  2022{\natexlab{a}}.
\newblock \href {https://arxiv.org/abs/2204.02311} {Palm: Scaling language modeling with pathways}.
\newblock \emph{arxiv:2204.02311}.

\bibitem[{Chowdhery et~al.(2022{\natexlab{b}})Chowdhery, Narang, Devlin, Bosma, Mishra, Roberts, Barham, Chung, Sutton, Gehrmann et~al.}]{chowdhery2022palm}
Aakanksha Chowdhery, Sharan Narang, Jacob Devlin, Maarten Bosma, Gaurav Mishra, Adam Roberts, Paul Barham, Hyung~Won Chung, Charles Sutton, Sebastian Gehrmann, et~al. 2022{\natexlab{b}}.
\newblock Palm: Scaling language modeling with pathways.
\newblock \emph{arXiv preprint arXiv:2204.02311}.

\bibitem[{Cobbe et~al.(2021)Cobbe, Kosaraju, Bavarian, Chen, Jun, Kaiser, Plappert, Tworek, Hilton, Nakano et~al.}]{cobbe2021training}
Karl Cobbe, Vineet Kosaraju, Mohammad Bavarian, Mark Chen, Heewoo Jun, Lukasz Kaiser, Matthias Plappert, Jerry Tworek, Jacob Hilton, Reiichiro Nakano, et~al. 2021.
\newblock Training verifiers to solve math word problems.
\newblock \emph{arXiv preprint arXiv:2110.14168}.

\bibitem[{Feng and Karmaker(2023)}]{feng2023joint}
Dongji Feng and Shubhra~Kanti Karmaker. 2023.
\newblock Joint upper \& expected value normalization for evaluation of retrieval systems: A case study with learning-to-rank methods.
\newblock \emph{Information Processing \& Management}, 60(4):103404.

\bibitem[{Han et~al.(2021)Han, Rute, Wu, Ayers, and Polu}]{han2021proof}
Jesse~Michael Han, Jason Rute, Yuhuai Wu, Edward~W Ayers, and Stanislas Polu. 2021.
\newblock Proof artifact co-training for theorem proving with language models.
\newblock \emph{arXiv preprint arXiv:2102.06203}.

\bibitem[{Kalyan et~al.(2021)Kalyan, Kumar, Chandrasekaran, Sabharwal, and Clark}]{kalyan2021much}
Ashwin Kalyan, Abhinav Kumar, Arjun Chandrasekaran, Ashish Sabharwal, and Peter Clark. 2021.
\newblock How much coffee was consumed during emnlp 2019? fermi problems: A new reasoning challenge for ai.
\newblock \emph{arXiv preprint arXiv:2110.14207}.

\bibitem[{Kojima et~al.(2022)Kojima, Gu, Reid, Matsuo, and Iwasawa}]{kojima2022large}
Takeshi Kojima, Shixiang~Shane Gu, Machel Reid, Yutaka Matsuo, and Yusuke Iwasawa. 2022.
\newblock Large language models are zero-shot reasoners.
\newblock \emph{arXiv preprint arXiv:2205.11916}.

\bibitem[{Li et~al.(2024)Li, Wen, Wang, Li, Yuan, Liu, Liu, Xu, Wang, Sun et~al.}]{li2024personal}
Yuanchun Li, Hao Wen, Weijun Wang, Xiangyu Li, Yizhen Yuan, Guohong Liu, Jiacheng Liu, Wenxing Xu, Xiang Wang, Yi~Sun, et~al. 2024.
\newblock Personal llm agents: Insights and survey about the capability, efficiency and security.
\newblock \emph{arXiv preprint arXiv:2401.05459}.

\bibitem[{Liu et~al.(2023)Liu, Yuan, Fu, Jiang, Hayashi, and Neubig}]{liu2023pre}
Pengfei Liu, Weizhe Yuan, Jinlan Fu, Zhengbao Jiang, Hiroaki Hayashi, and Graham Neubig. 2023.
\newblock Pre-train, prompt, and predict: A systematic survey of prompting methods in natural language processing.
\newblock \emph{ACM Computing Surveys}, 55(9):1--35.

\bibitem[{Lu et~al.(2022)Lu, Qiu, Yu, Welleck, and Chang}]{lu2022survey}
Pan Lu, Liang Qiu, Wenhao Yu, Sean Welleck, and Kai-Wei Chang. 2022.
\newblock A survey of deep learning for mathematical reasoning.
\newblock \emph{arXiv preprint arXiv:2212.10535}.

\bibitem[{Lu et~al.(2024)Lu, Zhou, Ren, Wang, Shi, Pan, Zhan, and Li}]{lu2024mathgenie}
Zimu Lu, Aojun Zhou, Houxing Ren, Ke~Wang, Weikang Shi, Junting Pan, Mingjie Zhan, and Hongsheng Li. 2024.
\newblock Mathgenie: Generating synthetic data with question back-translation for enhancing mathematical reasoning of llms.
\newblock \emph{arXiv preprint arXiv:2402.16352}.

\bibitem[{Manktelow(2012)}]{manktelow2012thinking}
Ken Manktelow. 2012.
\newblock \emph{Thinking and reasoning: An introduction to the psychology of reason, judgment and decision making}.
\newblock Psychology Press.

\bibitem[{Polu et~al.(2022)Polu, Han, Zheng, Baksys, Babuschkin, and Sutskever}]{polu2022formal}
Stanislas Polu, Jesse~Michael Han, Kunhao Zheng, Mantas Baksys, Igor Babuschkin, and Ilya Sutskever. 2022.
\newblock Formal mathematics statement curriculum learning.
\newblock \emph{arXiv preprint arXiv:2202.01344}.

\bibitem[{Polu and Sutskever(2020)}]{polu2020generative}
Stanislas Polu and Ilya Sutskever. 2020.
\newblock Generative language modeling for automated theorem proving.
\newblock \emph{arXiv preprint arXiv:2009.03393}.

\bibitem[{Radford et~al.(2019)Radford, Wu, Child, Luan, Amodei, Sutskever et~al.}]{radford2019language}
Alec Radford, Jeffrey Wu, Rewon Child, David Luan, Dario Amodei, Ilya Sutskever, et~al. 2019.
\newblock Language models are unsupervised multitask learners.
\newblock \emph{OpenAI blog}, 1(8):9.

\bibitem[{Rahgouy et~al.(2023)Rahgouy, Giglou, Feng, Rahgooy, Dozier, and Seals}]{rahgouy2023navigating}
Mostafa Rahgouy, Hamed~Babaei Giglou, Dongji Feng, Taher Rahgooy, Gerry~V Dozier, and Cheryl~D Seals. 2023.
\newblock Navigating the fermi multiverse: Assessing llms for complex multi-hop queries.
\newblock In \emph{NL4AI@ AI* IA}.

\bibitem[{Salvador et~al.(2024)Salvador, Bansal, Akter, Sarkar, Das, and Karmaker}]{salvador2024benchmarking}
John Salvador, Naman Bansal, Mousumi Akter, Souvika Sarkar, Anupam Das, and Shubhra~Kanti Karmaker. 2024.
\newblock Benchmarking llms on the semantic overlap summarization task.
\newblock \emph{arXiv preprint arXiv:2402.17008}.

\bibitem[{Santu and Feng(2023)}]{santu2023teler}
Shubhra Kanti~Karmaker Santu and Dongji Feng. 2023.
\newblock \href {http://arxiv.org/abs/2305.11430} {Teler: A general taxonomy of llm prompts for benchmarking complex tasks}.

\bibitem[{Santu et~al.(2024)Santu, Sinha, Bansal, Knipper, Sarkar, Salvador, Mahajan, Guttikonda, Akter, Freestone et~al.}]{santu2024prompting}
Shubhra Kanti~Karmaker Santu, Sanjeev~Kumar Sinha, Naman Bansal, Alex Knipper, Souvika Sarkar, John Salvador, Yash Mahajan, Sri Guttikonda, Mousumi Akter, Matthew Freestone, et~al. 2024.
\newblock Prompting llms to compose meta-review drafts from peer-review narratives of scholarly manuscripts.
\newblock \emph{arXiv preprint arXiv:2402.15589}.

\bibitem[{Scao et~al.(2022)Scao, Fan, Akiki, Pavlick, Ili{\'c}, Hesslow, Castagn{\'e}, Luccioni, Yvon, Gall{\'e} et~al.}]{scao2022bloom}
Teven~Le Scao, Angela Fan, Christopher Akiki, Ellie Pavlick, Suzana Ili{\'c}, Daniel Hesslow, Roman Castagn{\'e}, Alexandra~Sasha Luccioni, Fran{\c{c}}ois Yvon, Matthias Gall{\'e}, et~al. 2022.
\newblock Bloom: A 176b-parameter open-access multilingual language model.
\newblock \emph{arXiv preprint arXiv:2211.05100}.

\bibitem[{Tong et~al.(2024)Tong, Li, Wang, Wang, Teng, and Shang}]{tong2024can}
Yongqi Tong, Dawei Li, Sizhe Wang, Yujia Wang, Fei Teng, and Jingbo Shang. 2024.
\newblock Can llms learn from previous mistakes? investigating llms' errors to boost for reasoning.
\newblock \emph{arXiv preprint arXiv:2403.20046}.

\bibitem[{Touvron et~al.(2023)Touvron, Lavril, Izacard, Martinet, Lachaux, Lacroix, Rozi{\`e}re, Goyal, Hambro, Azhar et~al.}]{touvron2023llama}
Hugo Touvron, Thibaut Lavril, Gautier Izacard, Xavier Martinet, Marie-Anne Lachaux, Timoth{\'e}e Lacroix, Baptiste Rozi{\`e}re, Naman Goyal, Eric Hambro, Faisal Azhar, et~al. 2023.
\newblock Llama: Open and efficient foundation language models.
\newblock \emph{arXiv preprint arXiv:2302.13971}.

\bibitem[{Vaswani(2017)}]{vaswani2017attention}
Ashish Vaswani. 2017.
\newblock Attention is all you need.
\newblock \emph{arXiv preprint arXiv:1706.03762}.

\bibitem[{Wang et~al.(2022)Wang, Wei, Schuurmans, Le, Chi, Narang, Chowdhery, and Zhou}]{wang2022self}
Xuezhi Wang, Jason Wei, Dale Schuurmans, Quoc Le, Ed~Chi, Sharan Narang, Aakanksha Chowdhery, and Denny Zhou. 2022.
\newblock Self-consistency improves chain of thought reasoning in language models.
\newblock \emph{arXiv preprint arXiv:2203.11171}.

\bibitem[{Wei et~al.(2022)Wei, Wang, Schuurmans, Bosma, Chi, Le, and Zhou}]{wei2022chain}
Jason Wei, Xuezhi Wang, Dale Schuurmans, Maarten Bosma, Ed~Chi, Quoc Le, and Denny Zhou. 2022.
\newblock Chain of thought prompting elicits reasoning in large language models.
\newblock \emph{arXiv preprint arXiv:2201.11903}.

\bibitem[{Wei(2022)}]{wei-2022-entropy}
Yuxiang Wei. 2022.
\newblock \href {https://aclanthology.org/2022.amta-wetpr.8} {Entropy as a measurement of cognitive load in translation}.
\newblock In \emph{Proceedings of the 15th biennial conference of the Association for Machine Translation in the Americas (Workshop 1: Empirical Translation Process Research)}, pages 75--86. Association for Machine Translation in the Americas.

\bibitem[{Xia et~al.(2024)Xia, Kim, Chen, Ye, Kundu, Talati et~al.}]{xia2024understanding}
Yuchen Xia, Jiho Kim, Yuhan Chen, Haojie Ye, Souvik Kundu, Nishil Talati, et~al. 2024.
\newblock Understanding the performance and estimating the cost of llm fine-tuning.
\newblock \emph{arXiv preprint arXiv:2408.04693}.

\bibitem[{Yang et~al.(2024)Yang, Gribovskaya, Kassner, Geva, and Riedel}]{yang2024large}
Sohee Yang, Elena Gribovskaya, Nora Kassner, Mor Geva, and Sebastian Riedel. 2024.
\newblock Do large language models latently perform multi-hop reasoning?
\newblock \emph{arXiv preprint arXiv:2402.16837}.

\bibitem[{Yao et~al.(2024)Yao, Yu, Zhao, Shafran, Griffiths, Cao, and Narasimhan}]{yao2024tree}
Shunyu Yao, Dian Yu, Jeffrey Zhao, Izhak Shafran, Tom Griffiths, Yuan Cao, and Karthik Narasimhan. 2024.
\newblock Tree of thoughts: Deliberate problem solving with large language models.
\newblock \emph{Advances in Neural Information Processing Systems}, 36.

\bibitem[{Yao et~al.(2021)Yao, Huang, Wang, Dong, and Wei}]{yao-etal-2021-adapt}
Yunzhi Yao, Shaohan Huang, Wenhui Wang, Li~Dong, and Furu Wei. 2021.
\newblock \href {https://doi.org/10.18653/v1/2021.findings-acl.40} {Adapt-and-distill: Developing small, fast and effective pretrained language models for domains}.
\newblock In \emph{Findings of the Association for Computational Linguistics: ACL-IJCNLP 2021}, pages 460--470, Online. Association for Computational Linguistics.

\bibitem[{Yin et~al.(2024)Yin, Sun, Guo, Zeng, Li, Dai, Cheng, Huang, and Qiu}]{yin2024reasoning}
Zhangyue Yin, Qiushi Sun, Qipeng Guo, Zhiyuan Zeng, Xiaonan Li, Junqi Dai, Qinyuan Cheng, Xuan-Jing Huang, and Xipeng Qiu. 2024.
\newblock Reasoning in flux: Enhancing large language models reasoning through uncertainty-aware adaptive guidance.
\newblock In \emph{Proceedings of the 62nd Annual Meeting of the Association for Computational Linguistics (Volume 1: Long Papers)}, pages 2401--2416.

\bibitem[{Yoran et~al.(2023{\natexlab{a}})Yoran, Wolfson, Bogin, Katz, Deutch, and Berant}]{yoran2023answering}
Ori Yoran, Tomer Wolfson, Ben Bogin, Uri Katz, Daniel Deutch, and Jonathan Berant. 2023{\natexlab{a}}.
\newblock Answering questions by meta-reasoning over multiple chains of thought.
\newblock \emph{arXiv preprint arXiv:2304.13007}.

\bibitem[{Yoran et~al.(2023{\natexlab{b}})Yoran, Wolfson, Ram, and Berant}]{yoran2023making}
Ori Yoran, Tomer Wolfson, Ori Ram, and Jonathan Berant. 2023{\natexlab{b}}.
\newblock Making retrieval-augmented language models robust to irrelevant context.
\newblock \emph{arXiv preprint arXiv:2310.01558}.

\bibitem[{Zhang et~al.(2024)Zhang, Liu, Cherry, and Firat}]{zhang2024scaling}
Biao Zhang, Zhongtao Liu, Colin Cherry, and Orhan Firat. 2024.
\newblock When scaling meets llm finetuning: The effect of data, model and finetuning method.
\newblock \emph{arXiv preprint arXiv:2402.17193}.

\bibitem[{Zhao et~al.(2020)Zhao, Shang, Liu, Wang, and Liu}]{zhao2020ape210k}
Wei Zhao, Mingyue Shang, Yang Liu, Liang Wang, and Jingming Liu. 2020.
\newblock Ape210k: A large-scale and template-rich dataset of math word problems.
\newblock \emph{arXiv preprint arXiv:2009.11506}.

\bibitem[{Zhou et~al.(2022)Zhou, Sch{\"a}rli, Hou, Wei, Scales, Wang, Schuurmans, Bousquet, Le, and Chi}]{zhou2022least}
Denny Zhou, Nathanael Sch{\"a}rli, Le~Hou, Jason Wei, Nathan Scales, Xuezhi Wang, Dale Schuurmans, Olivier Bousquet, Quoc Le, and Ed~Chi. 2022.
\newblock Least-to-most prompting enables complex reasoning in large language models.
\newblock \emph{arXiv preprint arXiv:2205.10625}.

\end{thebibliography}
\bibliographystyle{acl_natbib}

\section{Ethics Statement}

In this paper, we have discussed about behavior of LLMs whiling prompting by TELeR taxonomy. Through this, we hope to assist new research direction. To the fulfilment of this goal, we have worked with real-world dataset. We did not obtain any explicit approval as our intended contents were already published for educational/research purposes. We have not tried to identify any private information from the data in any way which can result in a privacy violation. Additionally, the data we used (publicly released) does not contain personal information (e.g., usernames of users). In the whole experiment, we only used open source packages and libraries, along with proper citations as required also in accordance with its acceptable use policy, and no additional permission was required.

\appendix

\section{Appendix}\label{sec:appendix}

\subsection{TELeR Taxonomy}

\begin{figure*}[!htb]
    \centering

    \includegraphics[width=0.89\linewidth]{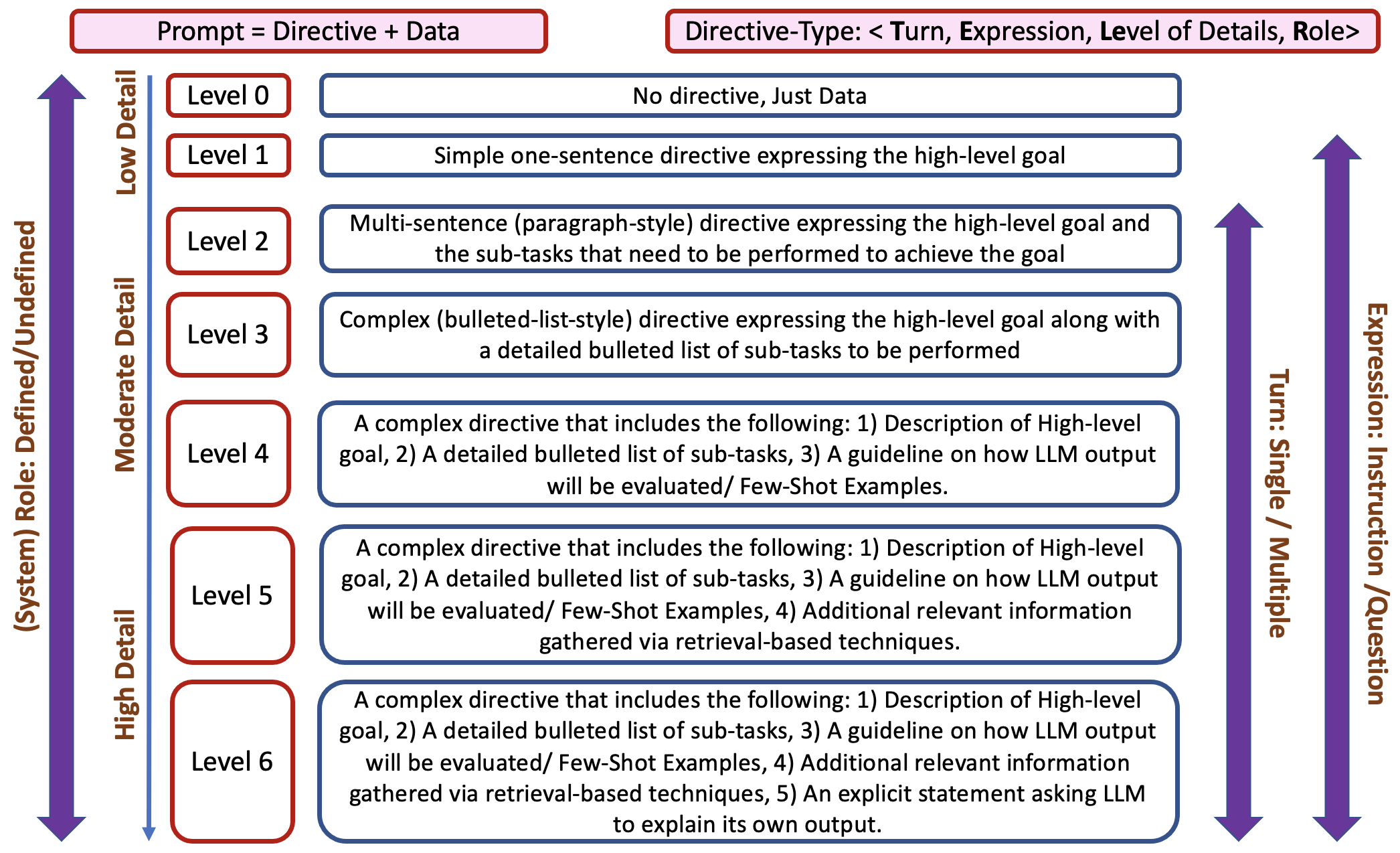}
    \vspace{-4mm}
    \caption{Proposed Prompt Taxonomy: \textbf{TELeR} (<\textbf{T}urn, \textbf{E}xpression, \textbf{Le}vel of Details, \textbf{R}ole>). This figure provides us a blueprint, the original sample to design prompts level. We selected its Level 0, Level 2, Level 4 to test its stimulation performance on LLM Models.~\cite{santu2023teler} }
    \label{fig:prompts_taxonomy}
    \vspace{-2mm}
\end{figure*}

Figure~\ref{fig:prompts_taxonomy} provides a blueprint TELeR taxonomy to design prompt level.

\onecolumn
\subsection{Prompt Examples for LLMs to classify FPs into standard or specific}
\label{sec:template2}

\begin{tcolorbox}[colback=lightgray!10, colframe=black, title=Prompts format for LLMs to categorize FPs]
\fontsize{8pt}{9.8pt}
\begin{lstlisting}[breaklines=true, basicstyle=\ttfamily, frame=none]
###role: 
"system", "content": "You are a helpful assistant that classifies questions as 'standard'or 'nonstandard'.""role":"user","content":"""

###messages:
The following question needs to be classified as 'standard' or 'nonstandard' based on the following criteria:
- Clarity: The question should have one intended meaning and be easy to understand without any ambiguity.
-Neutrality: The question should not be biased towards any particular viewpoint.
-Conciseness: The question should be concise and to the point.

###Examples:
"How much energy does it take to repair damaged tissue?" -> standard

"How many golf balls put into the world's oceans would it take to submere all of the land on earth from the displaced water?"-> nonstandard

###Output:
questions: {question}
Based on the criteria, this question is:"""

\end{lstlisting}
\end{tcolorbox}




\onecolumn
\subsection{Prompt Examples for Level 2 and Level 4 for one same questions}
\label{sec:template3}

\begin{tcolorbox}[colback=lightgray!10, colframe=black, title=Question: How many golf balls put into the worlds oceans would it take to submerge all of the land on
earth from the displaced water?]
\fontsize{8pt}{9.8pt}
\begin{lstlisting}[breaklines=true, basicstyle=\ttfamily, frame=none]
###Prompt2: 

High-level Goal: Calculate the number of golf balls needed to submerge all of
the land on Earth under the displaced water. 

Sub-tasks:
1. Research the average volume of a golf ball.
2. Determine the average density of a golf ball.
3. Calculate the total volume of land on Earth that needs to be submerged.
4. Calculate the volume of water displaced by submerging the land.
5. Divide the displaced water volume by the volume of a golf ball to find the total number of golf balls needed.
6. Consider factors like compression of golf balls and variations in land height for more accurate calculations.
7. Present the final result with appropriate units and context to convey the magnitude of the number of golf balls
required.

###Prompt4:

High-level Goal: Estimate the number of golf balls needed to submerge all the
land on Earth by calculating the volume of water displaced by them.

Sub-tasks:
1. Research the average volume of a golf ball.
2. Determine the average density of a golf ball.
3. Calculate the total volume of land on Earth that needs to be submerged.
4. Calculate the volume of water displaced by submerging the land.
5. Divide the displaced water volume by the volume of a golf ball to find the total number of golf balls needed.
6. Consider factors like compression of golf balls and variations in land height for more accurate calculations.
7. Present the final result with appropriate units and context to convey the magnitude of the number of golf balls
required.

Evaluation:- Accuracy of calculations- Logical progression of steps- Clarity in
presenting the final estimated number of golf balls.

Examples:1. Calculate the number of golf balls required to submerge all the land on
Earth if the average golf ball volume is 40 cubic centimeters.2. Estimate the number
of golf balls needed to submerge all the land on Earth assuming each golf ball
displaces 20 cubic centimeters of water.
\end{lstlisting}
\end{tcolorbox}




\end{document}